\documentclass[conference]{IEEEtran}
\IEEEoverridecommandlockouts

\usepackage{cite}
\usepackage{amsmath,amssymb}
\usepackage{graphicx}
\usepackage{booktabs}
\usepackage{xcolor}
\usepackage{enumitem}
\usepackage[hidelinks]{hyperref}

\setlist[itemize]{leftmargin=1.1em,itemsep=1pt,topsep=2pt}

\begin{document}

\title{RoboFake: Are Deepfake Detectors Ready for Humanoid Face Robots?}

\author{\IEEEauthorblockN{Hao Chen}
\IEEEauthorblockA{haoc.research@gmail.com}
\and 
\IEEEauthorblockN{Junnan Xu}
\IEEEauthorblockA{junnanxu25@gmail.com}}

\maketitle

\begin{abstract}
Humanoid face robots can now move their lips in sync with speech and show
detailed facial expressions. Give such a robot the face of a real person and
drive it with generated speech, and you get a new kind of presentation attack:
a face that talks and reacts in front of the camera, recorded as a real object
with no edited pixel. We call this a \emph{RoboFake} and ask whether current
defenses are ready for it. We test three public face presentation attack
detection (PAD) models and six public deepfake detectors on video of a physical
lip-syncing face robot, on a synthesized video of the same robot, and on
rendered robot faces. Thresholds are fixed on human data (BPCER 5\% on
FaceForensics++), and every result is compared with real human faces processed
the same way as the robot videos and turned to the same head angle. A small
mobile PAD model detects every physical robot frame (APCER 0\%) and is not affected by
compression, but its BPCER on processed human video rises to 14\%; the two PAD
models trained on CelebA-Spoof miss the robot in the direct capture, and one
of them also misses most panel frames. Among the deepfake detectors, three flag almost every robot frame and
two fewer than 7\%; masking shows that the response comes from the face but
does not track synthesis, and JPEG compression pushes most robot frames
below the threshold for some detectors. Our results suggest that current models
catch today's robots because their faces do not look human, not because they
recognize the attack. We state three testable hypotheses about realistic,
interactive RoboFakes.
\end{abstract}

\section{Introduction}
\label{sec:intro}

Humanoid face robots have improved quickly in the last few years. They can now
predict and mirror a person's facial expression~\cite{hu2024coexpression}, move
their lips in sync with speech in languages they were never trained
on~\cite{hu2026lip}, and learn expression control from large datasets of robot
faces~\cite{li2025x2c,zhu2025rena}. Androids built to look human, in some cases
like a specific person, have existed for even
longer~\cite{sakamoto2007android,sato2022nikola}. The faces of most of these
robots are still clearly not human. But the parts that make a face believable,
such as skin, lip motion and expression timing, are each being worked on.

Put these parts together and you get a new way to fake a person. Give a robot
the face of a real person, drive it with a cloned voice and a lip-sync model,
and place it in front of a camera, for example in a video call, or in front of
people at a front desk. We call this a \emph{RoboFake}. In biometric terms it
is a presentation attack, but an unusual one: the attack instrument is a face
that moves, talks and reacts, not a photo, a screen or a static mask. Unlike a
deepfake, nothing in the resulting video is edited: the camera records a real
object under real light, and the fake happens before the camera, not after it.
The generative models behind today's deepfakes, such as voice cloning and
audio-driven facial motion, now drive a physical face instead of producing
pixels. The targets are those of face recognition and liveness checks: video
calls, remote identity verification, and people who simply trust what they
see.

It is not clear that existing defenses cover this case. Deepfake detectors
look for traces of digital editing, such as blending boundaries, frequency
artifacts or generator fingerprints~\cite{rossler2019ffpp,qian2020f3net,shiohara2022sbi},
and a RoboFake has none of these. Face presentation attack detection (PAD)
deals with photos, screens and masks held up to a
camera~\cite{george2019wmca,liu2021hifimask,yu2023fassurvey}, but not with a
fake head that talks and reacts. Work that tells humanoids from humans looks
at body motion, not faces~\cite{kim2026irobot,li2026motionturing}. And defenses
against live deepfakes rely on the generated face breaking when the scene
changes~\cite{mittal2024gotcha,xie2024sfake}, which a physical face does not do.

In this paper we take a first step and ask what current PAD models and
deepfake detectors do when they see a robot face. We do not build a RoboFake. We use public data of a physical
lip-syncing face robot~\cite{hu2026lip}, which includes a synthesized video of
the same robot speaking the same sentences, plus rendered robot faces from
X2C~\cite{li2025x2c}. We test three public PAD models and six public deepfake
detectors. All thresholds are fixed on human data, and every robot result is
compared with real human faces that went through the same processing as the
robot videos. Our main findings are:
\begin{itemize}
\item \textbf{Deepfake detectors disagree.} At a threshold that flags 5\% of
      real faces, F3Net, SRM and Effort flag 97--100\% of the robot frames,
      Xception 47\%, and EfficientNet-B4 and SBI fewer than 7\%.
\item \textbf{The response comes from the face but does not track synthesis.}
      With only the background visible, detectors no longer score the robot
      higher than humans (AUC $\le$0.61). With only the face visible, three of six
      give the physical robot a higher fake score than its synthesized video.
\item \textbf{The response is fragile.} JPEG compression shifts the
      threshold enough that Xception drops from 47\% to 9\% of robot frames
      flagged at quality 90, and Effort from 100\% to 33\% at quality 75.
\item \textbf{Liveness models do better, but not reliably.} A small mobile PAD
      model flags every physical robot frame, survives compression and runs in
      under 6\,ms on a CPU, but it also flags 14\% of processed human frames.
      AENet flags the robot panels but misses the direct capture, and MN3
      misses most of both.
\end{itemize}
Together, these results suggest that current detectors catch today's robots
because their faces do not look human, not because they recognize a forgery.

Our contributions are: (1) we define RoboFake as a presentation attack with a
moving, talking face and give a threat model that separates it from deepfakes,
classic presentation attacks and body-level humanoid detection; (2) we propose an evaluation protocol for physical face forgeries,
with thresholds fixed on human data, human controls matched in processing and
head angle, and errors reported as APCER and BPCER, and
will release our code; (3) we report the first measurements of deepfake
detectors and face PAD models on physical robot faces, a forgery source none of
them has seen, including tests that separate the face from the background and
the physical robot from its synthesized copy, and tests under compression; and (4) we state three testable hypotheses about realistic,
interactive RoboFakes.

\section{Related Work}
\label{sec:related}

\paragraph{Face forgery detection.}
Most face forgery detectors are trained to tell real faces from digitally
manipulated ones. FaceForensics++ (FF++)~\cite{rossler2019ffpp} made this a
standard task, with an Xception classifier~\cite{chollet2017xception} as a
strong baseline. Later work looks for cues that carry over to unseen
manipulations: frequency and noise-residual inputs (F3Net~\cite{qian2020f3net},
SRM~\cite{luo2021srm}), self-blended training images (SBI~\cite{shiohara2022sbi}),
and large pretrained backbones (Effort~\cite{yan2025effort}, built on
CLIP~\cite{radford2021clip}). DeepfakeBench~\cite{yan2023deepfakebench} puts many
of these under one codebase, which we use. Other methods use mouth
motion~\cite{haliassos2021lips} or blood-flow signals~\cite{ciftci2020fakecatcher}.
A recent study shows that current detectors flag many real smartphone photos as
fake~\cite{laionmobile2026}, which is why we pair every robot result with a
human control processed the same way. All of this work assumes the forgery is
in the pixels.

\paragraph{Humanoid face robots.}
Mori's uncanny valley~\cite{mori2012uncanny} describes how near-human faces can
feel unsettling, and later studies measured this on real robot
faces~\cite{mathur2016navigating}. Some androids copy a specific person:
Geminoid was built as a copy of its designer and used as a telepresence
medium~\cite{sakamoto2007android}. Recent work learns to control robot faces:
Emo predicts and mirrors expressions~\cite{hu2024coexpression}, Hu et al.\ learn
lip motion from speech audio~\cite{hu2026lip}, and X2C~\cite{li2025x2c} and
Rena~\cite{zhu2025rena} provide data and models for robot expressions. This line
of work measures realism as seen by people; we look at how it is seen by
detectors.

\paragraph{Telling humanoids from humans.}
I-Robot~\cite{kim2026irobot} and the Motion Turing Test~\cite{li2026motionturing}
separate humanoid from human body motion; neither looks at faces.

\paragraph{Presentation attacks and live deepfakes.}
Face PAD protects face recognition against printed photos, screen replays and
masks~\cite{yu2023fassurvey}. Most mask attacks, as in
HiFiMask~\cite{liu2021hifimask}, are worn by a live person. Some datasets also
include static fake heads: WMCA~\cite{george2019wmca} contains mannequin heads,
some warmed before capture to fool thermal sensors, and SiW-M~\cite{liu2019siwm}
contains mannequin attacks. These heads do not move, speak or respond. A face
robot is, in this sense, a fake head that does all three, and to our knowledge
no PAD study has tested one. We test three public PAD models, two of them
trained on CelebA-Spoof~\cite{zhang2020celebaspoof}. For live deepfakes in video
calls, GOTCHA~\cite{mittal2024gotcha} asks the caller to perform actions that
are hard for real-time generators, and SFake~\cite{xie2024sfake} makes the phone
vibrate and checks whether the blur on the face matches. Both rely on the
generated face failing when the physical scene changes.

\section{RoboFake: Definition and Threat Model}
\label{sec:threat}

We define a \emph{RoboFake} as a physical humanoid face robot that shows a face
to a camera or to people, with the aim of being taken for a real human, either
a specific person or a person in general. Its face can be shaped and colored to
look like someone, and its speech and expressions can be driven by software,
for example a voice clone and a lip-sync controller such as~\cite{hu2026lip}.
Unlike a photo, screen or mask, it moves, talks and reacts by itself, so
liveness checks based on motion or responses may not stop it. It differs from a
deepfake in three ways.
\emph{Physical capture:} the camera films a real object, so lighting, shadows
and camera motion are physically consistent, and there are no blending
boundaries, generator fingerprints or editing flicker. Any signal a deepfake
detector picks up must come from somewhere else, such as the look of the face.
\emph{Real-time interaction:} a robot can be connected to a dialogue system and
react to people. A live deepfake can too, but it has to render every frame and
may fail under unexpected conditions~\cite{mittal2024gotcha}; a robot renders
nothing. \emph{Physical presence:} a RoboFake can be met in person, where there
may be no recorded video at all, only the defender's own camera.

\paragraph{Attacker and defender.}
The attacker controls the robot, its face, voice and behavior. The defender sees
only a video stream, possibly compressed, for example in a video call or a
remote identity-verification session, and runs an off-the-shelf deepfake
detector or face PAD model. The defender does not know in advance that a robot
might be involved and does not interact with it on purpose. We study this
passive setting; active defenses are discussed in Sec.~\ref{sec:disc}.

\begin{figure}[t]
\centering
\includegraphics[width=\columnwidth]{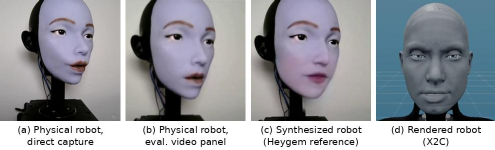}
\caption{Robot faces used in this study. (a) Direct capture of the physical
robot. (b) A physical-robot panel from an evaluation video. (c) The synthesized
reference panel of the same robot. (d) A rendered robot face from X2C. Panels
(b) and (c) come from the same composite video and show the same sentence.}
\label{fig:samples}
\end{figure}

\paragraph{Scope.}
The robot faces available today do not look human
(Fig.~\ref{fig:samples}). So we cannot test whether a realistic RoboFake would
fool current detectors. We can test how detectors react to a robot face that is
clearly not human, and whether they flag it for a reason that will still hold
when robot faces become realistic. Our results should not be read as evidence
that robots are easy to catch.

\section{Data and Evaluation Protocol}
\label{sec:protocol}

We only use public data and public checkpoints, and retrain nothing.
Table~\ref{tab:data} lists all test groups.

\paragraph{Robot data.}
Our main source is the data released with the lip-sync study of Hu
et al.~\cite{hu2026lip,hu2026lipdata}. Their user study showed people composite
videos: the top panel is a target video of the robot synthesized with
Heygem~\cite{heygem}, a talking-face generator; the three panels below are
recordings of the physical robot speaking the same audio, driven by three
controllers (an amplitude baseline, a nearest-neighbor landmark baseline, and
the authors' method), with the head at about $30^\circ$ to the camera. There are
13 such videos, cut from three English sentences. We follow the paper's
supplement here; the dataset README calls these videos ``fully synthesized'',
which as far as we can tell refers to the composite as a whole. This gives the
same robot and sentence once as a synthesized video and three times as a real
recording. We cut each panel into its own clip (H.264, CRF 18), giving 39
physical and 13 synthesized clips. The release also has one 10-second direct
capture of the robot, which we use as a second capture. We add 5{,}000 rendered
faces of a different robot from X2C~\cite{li2025x2c}.

\paragraph{Human data and the same-pipeline control.}
We use the official FF++ c23 test split~\cite{rossler2019ffpp}: 140 real videos
and the matching videos of four manipulations. FF++ fixes the thresholds and
serves as a sanity check. FF++ alone is not a fair comparison, because the robot
clips went through different processing: the face fills about 66\% of the
panel, the panel was resized, and each clip was encoded three times (camera,
composite, our cropping). If this processing made faces look fake by itself, we
would wrongly blame the robot. So we push the 140 real FF++ test videos (already
H.264 at CRF 23) through the same steps: crop so the face fills 66\% of the
frame, resize to $432\times432$, and encode twice more, with CRF 23 for the
composite step whose settings we do not know (23 is the x264 default) and CRF
18 for our cropping. The robot also turns its head about 30$^\circ$ away from
the camera, while most FF++ faces are close to frontal. We therefore build a
second control with the robot's head angle: we sample 160 frames per
same-pipeline video, estimate yaw from the five YuNet landmarks, and keep
frames inside the robot's 10th--90th percentile range (27--33$^\circ$; median
29$^\circ$ for the robot vs.\ 9$^\circ$ for the full control), at most 32 per
video. This gives 404 frames from 39 videos. As a very different human source
we use 2{,}000 FFHQ images~\cite{karras2019stylegan}.

\begin{table}[t]
\centering
\caption{Test data. ``Face'' is the median side of the detected face box in
pixels. For confidence intervals we resample clips, except that the four
panels of one composite video share the sentence and audio and are resampled
together (13 units); for X2C we resample its 10 image subsets.}
\label{tab:data}
\setlength{\tabcolsep}{3.2pt}
\small
\resizebox{\columnwidth}{!}{%
\begin{tabular}{llrrr}
\toprule
Group & Content & Clips & Frames & Face \\
\midrule
FF++ real (test)       & real human        & 140          & 4,480  & 166 \\
FF++ fakes (4 types)   & deepfake          & $4\times140$ & 17,920 & 166 \\
FF++ real, same pipe.  & real human        & 140          & 4,479  & ${\sim}$285 \\
\quad yaw-matched       & real human        & 39           & 404    & ${\sim}$285 \\
FFHQ                   & real human        & --           & 2,000  & 380 \\
Robot, eval.\ panels   & physical robot    & 39           & 1,245  & 287 \\
Robot, direct capture  & physical robot    & 1            & 320    & 330 \\
Robot, reference panel & synthesized robot & 13           & 415    & 284 \\
X2C                    & rendered robot    & (10)         & 5,000  & 408 \\
\bottomrule
\end{tabular}}
\end{table}

\paragraph{Preprocessing.}
We detect faces with YuNet~\cite{wu2023yunet} on 32 evenly spaced frames per
video, take a square box around the face enlarged by 1.3 (1.2 for the robot
panels and the same-pipeline control, kept above 82\% of the frame height to
avoid burned-in text), and resize it to the detector input. The direct capture
is short (320 frames), so we use all of them. YuNet found a face in every robot frame; a few
clip-end frames could not be decoded. We
save PNG crops and JPEG copies at quality 90 and 75.

\paragraph{Deepfake detectors.}
We test six detectors with public FF++ c23 checkpoints that cover the main
design families: Xception~\cite{rossler2019ffpp,chollet2017xception} and
EfficientNet-B4~\cite{tan2019efficientnet} (plain CNNs), F3Net~\cite{qian2020f3net}
and SRM~\cite{luo2021srm} (extra frequency or noise stream),
SBI~\cite{shiohara2022sbi} (trained on real faces with self-blended fakes), and
Effort~\cite{yan2025effort} (CLIP ViT-L/14). We use the DeepfakeBench checkpoints
and code for the first four, the official SBI weights at $380\times380$, and the
official Effort checkpoint in the DeepfakeBench implementation. Our FF++
frame-level AUC for the four DeepfakeBench models (0.974, 0.951, 0.978, 0.962)
is within 0.015 of the values DeepfakeBench reports; SBI (0.873) and Effort
(0.921) are lower in-domain, with their lowest scores on NeuralTextures
(Table~\ref{tab:sanity}).

\begin{table*}[t]
\centering
\caption{Sanity check on the FF++ c23 test split (140 videos per class, 32 frames per video). AUC is computed against FF++ real frames. All detectors use public checkpoints trained on FF++ c23; we did not retrain anything. ``Mean'' is the average over the four manipulations (equal to the AUC against all fakes pooled, since the four sets have the same size); ``Rep.'' is the FF-c23 frame-level AUC reported by DeepfakeBench~\cite{yan2023deepfakebench}; ``Video AUC'' uses clip-level mean scores. SBI is trained on real FF++ faces only, with self-blended images as fakes~\cite{shiohara2022sbi}, which explains its lower in-domain numbers.}
\label{tab:sanity}
\begin{tabular}{lllcccccccc}
\toprule
 & & & \multicolumn{6}{c}{Frame-level AUC} & & \\
\cmidrule(lr){4-9}
Detector & Backbone & Input & DF & F2F & FS & NT & Mean & Rep. & Video AUC & EER \\
\midrule
Xception & Xception & 256 & 0.986 & 0.983 & 0.975 & 0.951 & 0.974 & 0.964 & 0.989 & 0.082 \\
EffNet-B4 & EfficientNet-B4 & 256 & 0.944 & 0.953 & 0.977 & 0.930 & 0.951 & 0.957 & 0.975 & 0.116 \\
F3Net & Xception, 2-stream & 256 & 0.984 & 0.982 & 0.988 & 0.958 & 0.978 & 0.964 & 0.992 & 0.070 \\
SRM & Xception, 2-stream & 256 & 0.974 & 0.965 & 0.972 & 0.938 & 0.962 & 0.958 & 0.987 & 0.094 \\
SBI & EfficientNet-B4 & 380 & 0.963 & 0.856 & 0.897 & 0.776 & 0.873 & -- & 0.908 & 0.216 \\
Effort & CLIP ViT-L/14 & 224 & 0.989 & 0.916 & 0.971 & 0.807 & 0.921 & -- & 0.945 & 0.148 \\
\bottomrule
\end{tabular}
\end{table*}

\paragraph{Face PAD models.}
We also test three PAD models with public weights:
MiniFASNet~\cite{silentface}, two tiny CNNs (0.43M parameters each) from an
open-source mobile system whose outputs are averaged; AENet, the official CelebA-Spoof
model~\cite{zhang2020celebaspoof}; and MN3, a MobileNetV3 trained on
CelebA-Spoof~\cite{prokofiev2020mn3}. We crop faces as each official
inference code does, on the same frames and face boxes: the face box at
$224\times224$ for AENet, the box widened by 10\% at $128\times128$ for MN3,
and regions 2.7 and 4 times the box at $80\times80$ for MiniFASNet. CelebA-Spoof consists
of JPEG images, and both models trained on it react strongly to lossless input:
on PNG crops, AENet flags 61\% of FF++ real frames at its default threshold of
0.5, and so many real frames score exactly 1.0 that a 5\% FPR threshold cannot
be set. With JPEG 75 this drops to 3\%. We therefore use JPEG 75 input for
these two models. MiniFASNet
behaves the same on both formats. As a check of our pipeline, AENet and MN3
separate 2{,}000 live from 2{,}000 spoof CelebA-Spoof test images almost
perfectly (AUC $\ge$0.996); MiniFASNet, trained on other data, reaches 0.84.

\paragraph{Thresholds and metrics.}
We fix all thresholds on FF++ and never tune them on robot data: the EER point
between FF++ real and the four FF++ fakes pooled, and the point where 5\% of FF++
real frames are flagged (``5\% FPR''). A deployed detector is usually tuned to
keep false alarms on real people low, so 5\% FPR is our main operating point;
for PAD models it is the only one, since there are no FF++ attacks to
calibrate an EER. When the input format changes, we re-fit the thresholds on
FF++ in that format. In ISO/IEC 30107-3 terms~\cite{iso30107}, the 5\% FPR
threshold fixes the BPCER on FF++ real faces at 5\%; the share of frames
flagged is the BPCER on the other human groups and $1-$APCER on the robot
groups. Our main metric is the share of frames flagged in each group, with 95\% confidence intervals from resampling clips or composites
(2{,}000 rounds), since frames of one clip are not independent. We also report the AUC between
groups, which does not depend on any threshold.

\paragraph{Masking, Grad-CAM and timing.}
To see what drives the score, we gray out (value 128) either a fixed central
box covering 60\% of the width and 75\% of the height (``Bg.'' keeps the
background) or everything outside it (``Face''). For the CNN detectors we
compute Grad-CAM~\cite{selvaraju2017gradcam} on 128 frames per group. We time
each model on an AMD Ryzen 7 9700X CPU limited to 4 threads, at batch size 1,
excluding face detection.

\section{Results}
\label{sec:results}

\begin{table*}[t]
\centering
\caption{Percentage of frames flagged as fake (PNG input). Thresholds are set on FF++ c23 only and then frozen. ``Yaw-match.'': same-pipeline human frames with the robot's head angle (Sec.~\ref{sec:protocol}). Brackets: 95\% bootstrap intervals, rounded outward to whole percents. At 5\% FPR, FF++ real frames are flagged 5\% of the time by construction.
$^\dagger$SBI has a high FF++ EER (0.216) and its EER threshold is very low (0.044), so this row says little.
$^\ddagger$A single clip, so no interval.}
\label{tab:main}
\setlength{\tabcolsep}{4pt}
\begin{tabular}{lccccccc}
\toprule
 & \multicolumn{2}{c}{Human, same pipeline} & & \multicolumn{4}{c}{Robot} \\
\cmidrule(lr){2-3}\cmidrule(lr){5-8}
Detector & All & Yaw-match. & FFHQ & Physical (eval.) & Physical (direct)$^\ddagger$ & Synthesized & X2C \\
\midrule
\multicolumn{8}{l}{\emph{Threshold at FF++ EER}} \\
Xception & 15.9\,{\scriptsize[12,20]} & 18.3\,{\scriptsize[8,30]} & 18.3 & 94.6\,{\scriptsize[92,97]} & 100.0 & 99.3\,{\scriptsize[98,100]} & 99.5\,{\scriptsize[99,100]} \\
EffNet-B4 & 22.4\,{\scriptsize[18,27]} & 32.9\,{\scriptsize[18,48]} & 32.1 & 76.0\,{\scriptsize[71,81]} & 15.9 & 77.6\,{\scriptsize[73,82]} & 28.0\,{\scriptsize[21,37]} \\
F3Net & 19.9\,{\scriptsize[15,24]} & 27.0\,{\scriptsize[13,41]} & 20.4 & 99.5\,{\scriptsize[99,100]} & 100.0 & 91.6\,{\scriptsize[88,95]} & 83.9\,{\scriptsize[76,90]} \\
SRM & 19.6\,{\scriptsize[15,24]} & 32.7\,{\scriptsize[18,47]} & 16.6 & 100.0\,{\scriptsize[100,100]} & 100.0 & 100.0\,{\scriptsize[100,100]} & 100.0\,{\scriptsize[99,100]} \\
SBI$^\dagger$ & 24.4\,{\scriptsize[19,30]} & 30.7\,{\scriptsize[16,46]} & 73.3 & 98.2\,{\scriptsize[97,99]} & 100.0 & 100.0\,{\scriptsize[100,100]} & 99.8\,{\scriptsize[99,100]} \\
Effort & 31.4\,{\scriptsize[26,37]} & 29.7\,{\scriptsize[16,44]} & 12.0 & 100.0\,{\scriptsize[100,100]} & 100.0 & 100.0\,{\scriptsize[100,100]} & 100.0\,{\scriptsize[100,100]} \\
\midrule
\multicolumn{8}{l}{\emph{Threshold at 5\% FPR on FF++ real}} \\
Xception & 10.3\,{\scriptsize[7,14]} & 9.9\,{\scriptsize[2,19]} & 13.6 & 46.7\,{\scriptsize[42,52]} & 95.9 & 92.8\,{\scriptsize[89,96]} & 98.7\,{\scriptsize[98,100]} \\
EffNet-B4 & 11.0\,{\scriptsize[8,15]} & 17.3\,{\scriptsize[8,28]} & 19.9 & 6.4\,{\scriptsize[4,9]} & 0.0 & 21.2\,{\scriptsize[16,26]} & 15.8\,{\scriptsize[11,21]} \\
F3Net & 14.7\,{\scriptsize[11,19]} & 19.6\,{\scriptsize[9,32]} & 15.1 & 96.7\,{\scriptsize[95,98]} & 100.0 & 81.2\,{\scriptsize[75,87]} & 74.1\,{\scriptsize[64,82]} \\
SRM & 11.6\,{\scriptsize[8,15]} & 19.6\,{\scriptsize[9,31]} & 8.1 & 99.5\,{\scriptsize[99,100]} & 100.0 & 99.5\,{\scriptsize[98,100]} & 98.3\,{\scriptsize[97,99]} \\
SBI & 6.7\,{\scriptsize[4,10]} & 5.2\,{\scriptsize[1,10]} & 41.5 & 4.4\,{\scriptsize[2,7]} & 99.1 & 56.6\,{\scriptsize[48,65]} & 55.9\,{\scriptsize[49,65]} \\
Effort & 10.9\,{\scriptsize[7,15]} & 7.4\,{\scriptsize[2,14]} & 1.9 & 99.9\,{\scriptsize[99,100]} & 86.9 & 100.0\,{\scriptsize[100,100]} & 54.8\,{\scriptsize[52,58]} \\
\bottomrule
\end{tabular}
\end{table*}

\subsection{Do deepfake detectors flag the physical robot?}
\label{sec:res-flag}

At the EER threshold, five of six detectors flag at least 76\% of the physical
robot frames (SBI's EER threshold is only 0.044, so we do not use it). But this
threshold also flags 16--31\% of same-pipeline human frames. At 5\% FPR
(Table~\ref{tab:main}) the detectors split into three groups: F3Net, SRM and
Effort still flag almost all robot frames (96.7--99.9\%), Xception about half
(46.7\%), and EfficientNet-B4 and SBI very few (6.4\% and 4.4\%), below their
false-alarm rate on same-pipeline humans. Capture conditions matter too: SBI flags 4.4\% of the panel frames but 99.1\% of
the direct-capture frames, and EfficientNet-B4 goes the other way at the EER
threshold (76.0\% vs.\ 15.9\%). With a single direct clip, we take this only as
a warning.

\subsection{Is it the robot, the processing or the background?}
\label{sec:res-mask}

Same-pipeline human frames are flagged 6.7--14.7\% of the time at 5\% FPR,
above the nominal 5\% but far below the robot rate for four detectors. The
head angle does not explain the gap either: yaw-matched human frames are
flagged 5.2--19.6\%, a few points more than the full control for
EfficientNet-B4, F3Net and SRM, but still far below the 47--100\% that
Xception, F3Net, SRM and Effort give the robot. FFHQ
gives similar rates except for SBI (41.5\%), probably because of FFHQ's
alignment and blurred padding. The frame-level AUC between
same-pipeline humans and the robot is 0.81--0.95 for all six detectors
(Table~\ref{tab:mask}), so every detector scores the robot higher; for
EfficientNet-B4 and SBI the robot scores are simply still below an FF++
threshold. Masking shows where this comes from. With only the background, the
AUC falls to 0.61 or lower, and for Xception (0.12) and EfficientNet-B4 (0.14)
the robot's background even looks \emph{less} fake than a human one. With only
the face, the AUC stays at 0.84--0.95. The gray patch is out of distribution,
so for masked inputs we compare AUCs only.

\begin{table}[t]
\centering
\caption{Frame-level AUC under masking. Left: same-pipeline human (negative)
vs.\ physical robot. Right: physical robot (negative) vs.\ synthesized robot.
``Bg.'' keeps only the area outside a central face box; ``Face'' keeps only the
box. Below 0.5 the order is reversed.}
\label{tab:mask}
\setlength{\tabcolsep}{4pt}
\small
\begin{tabular}{lccccccc}
\toprule
 & \multicolumn{3}{c}{Human vs.\ robot} & & \multicolumn{3}{c}{Physical vs.\ synth.} \\
\cmidrule(lr){2-4}\cmidrule(lr){6-8}
Detector & Full & Bg. & Face & & Full & Bg. & Face \\
\midrule
Xception  & 0.89 & 0.12 & 0.89 & & 0.92 & 0.98 & 0.80 \\
EffNet-B4 & 0.81 & 0.14 & 0.86 & & 0.58 & 0.99 & 0.15 \\
F3Net     & 0.93 & 0.39 & 0.89 & & 0.43 & 0.98 & 0.32 \\
SRM       & 0.95 & 0.61 & 0.88 & & 0.95 & 1.00 & 0.38 \\
SBI       & 0.88 & 0.54 & 0.84 & & 0.85 & 0.10 & 0.67 \\
Effort    & 0.94 & 0.46 & 0.95 & & 1.00 & 0.26 & 1.00 \\
\bottomrule
\end{tabular}
\end{table}

\subsection{Physical robot vs.\ synthesized robot}
\label{sec:res-syn}

A deepfake detector is supposed to respond to synthesis, and each composite
video shows the same robot and sentence once synthesized and three times
filmed. On full frames, Xception, SRM, SBI and Effort score the synthesized
panel higher (AUC 0.85--1.00), F3Net and EfficientNet-B4 do not. But with only
the background, Xception, EfficientNet-B4, F3Net and SRM separate the two
kinds of panel almost perfectly (0.98--1.00) and SBI does so in reverse
(0.10), because the synthesized panel has its own background rendering.
With only the face, Effort (1.00), Xception (0.80) and SBI (0.67) still score
the synthesized face higher, while EfficientNet-B4 (0.15), F3Net (0.32) and SRM
(0.38) give the \emph{physical} face a higher fake score. For these three detectors, the fake score on a
robot face seems to measure how un-human the face looks, not whether the pixels
were generated.

\subsection{Effect of compression}
\label{sec:res-jpeg}

\begin{figure*}[t]
\centering
\includegraphics[width=\textwidth]{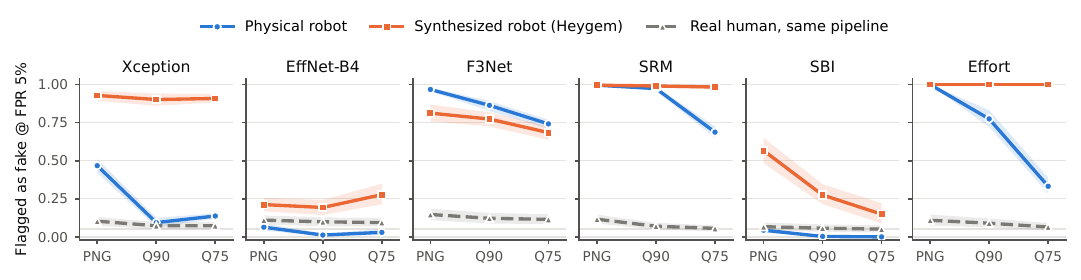}
\caption{Share of frames flagged at each detector's 5\% FPR threshold for PNG
and two JPEG qualities; the threshold is re-fit on FF++ for every format.
Bands: 95\% bootstrap intervals. Gray line: 5\%.}
\label{fig:compression}
\end{figure*}

Figure~\ref{fig:compression} repeats the evaluation on JPEG 90 and 75 copies.
Same-pipeline humans hardly move (5.1--14.7\% across detectors and formats),
but the physical robot drops a lot: Xception from 46.7\% to 9.4\% at quality 90,
Effort from 99.9\% to 77.4\% at 90 and 33.2\% at 75, SRM to 68.7\% and F3Net to
74.1\% at 75. The synthesized robot holds up better for most detectors. The ranking itself
survives: the human-vs.-robot AUC at quality 75 is still 0.77--0.95. What
changes is the threshold. Compressed FF++ real faces get higher scores, so the
5\% FPR threshold moves up (for Effort, from 0.70 to 0.85), while the mean robot
score stays about the same (0.83). The robot response is not large enough to
survive this shift.

\subsection{Where do the detectors look?}
\label{sec:res-cam}

\begin{figure*}[t]
\centering
\includegraphics[width=0.69\textwidth]{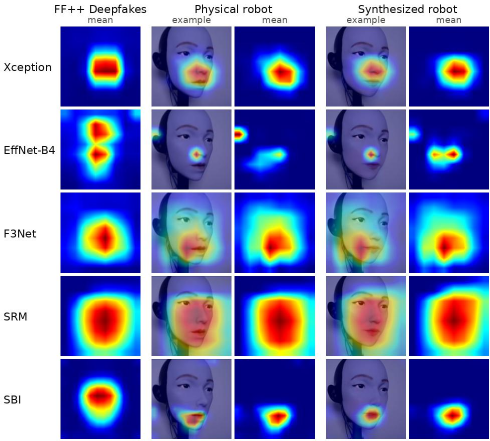}
\caption{Grad-CAM maps on the last convolution layer used in the forward pass:
one example and the mean map over 128 frames per group. For FF++ we show only
the mean map, so that no real person can be identified. Effort is a ViT and is
not included.}
\label{fig:gradcam}
\end{figure*}

\begin{table}[t]
\centering
\caption{Where the Grad-CAM mass falls (128 frames per group). ``Face'' is the share inside the central face box, ``Mouth'' the share inside the mouth box. The face box covers 45\% of the image area and the mouth box 10\%, so a uniform map would give 0.45 and 0.10. Effort is a ViT and is not included.}
\label{tab:gradcam}
\setlength{\tabcolsep}{3.5pt}
\resizebox{\columnwidth}{!}{%
\begin{tabular}{lcccccccc}
\toprule
 & \multicolumn{4}{c}{Face share} & & \multicolumn{3}{c}{Mouth share} \\
\cmidrule(lr){2-5}\cmidrule(lr){7-9}
Detector & DF & Robot & Synth. & X2C & & DF & Robot & Synth. \\
\midrule
Xception & 0.92 & 0.93 & 0.93 & 0.90 & & 0.22 & 0.31 & 0.28 \\
EffNet-B4 & 0.76 & 0.51 & 0.77 & 0.68 & & 0.16 & 0.20 & 0.23 \\
F3Net & 0.78 & 0.73 & 0.71 & 0.68 & & 0.24 & 0.22 & 0.23 \\
SRM & 0.71 & 0.70 & 0.69 & 0.79 & & 0.21 & 0.20 & 0.18 \\
SBI & 0.91 & 0.88 & 0.96 & 0.92 & & 0.21 & 0.71 & 0.74 \\
\bottomrule
\end{tabular}}
\end{table}

Fig.~\ref{fig:gradcam} and Table~\ref{tab:gradcam} show where the CNN detectors
look. Xception and SBI put 0.88--0.96 of the map mass inside the face box for
all groups. On the robot, SBI puts 71\% of the mass on the mouth box, against
21\% on FF++ Deepfakes; the mouth is the part the robot moves for lip sync and
the part a lip-sync generator changes most. F3Net and SRM spread their maps over
the whole face (0.70--0.73). EfficientNet-B4 puts only half of its mass on the
robot face and has a hot spot at the left image border, matching its weak and
unstable results.

\subsection{Do presentation attack detectors flag the robot?}
\label{sec:res-pad}

\begin{table}[t]
\centering
\caption{Face PAD models at the threshold that gives 5\% BPCER on FF++ real
frames. BPCER on the same-pipeline human control (all frames and yaw-matched),
APCER (\%) on the robot groups, AUC for same-pipeline human vs.\ robot (H/R)
and physical vs.\ synthesized robot (P/S), and CPU time per frame. MiniFASNet
time is the sum of its two models. Q75: JPEG 75.}
\label{tab:pad}
\setlength{\tabcolsep}{2.4pt}
\footnotesize
\resizebox{\columnwidth}{!}{%
\begin{tabular}{lccccccccc}
\toprule
 & \multicolumn{2}{c}{BPCER} & \multicolumn{4}{c}{APCER} & \multicolumn{2}{c}{AUC} & CPU \\
\cmidrule(lr){2-3}\cmidrule(lr){4-7}\cmidrule(lr){8-9}
Model & All & Yaw & Robot & Direct & Synth. & X2C & H/R & P/S & (ms) \\
\midrule
MiniFASNet (PNG) & 14.1 & 13.6 & 0   & 0   & 0    & 34.1 & 0.99 & 0.45 & 5.6 \\
MiniFASNet (Q75) & 13.6 & 16.6 & 0   & 0   & 0    & 5.8  & 0.97 & 0.84 & 5.6 \\
AENet (Q75)      & 4.9  & 5.7  & 1.8 & 100 & 74.9 & 100  & 0.98 & 0.02 & 8.4 \\
MN3 (Q75)        & 5.5  & 13.9 & 81.7 & 100 & 65.1 & 100 & 0.92 & 0.61 & 3.9 \\
\bottomrule
\end{tabular}}
\end{table}

All three PAD models rank the robot above same-pipeline humans (AUC 0.92--0.99,
Table~\ref{tab:pad}), but how many robot frames they flag differs a lot.
MiniFASNet flags every physical robot frame in both captures, and compression
does not change this. Its weak point is real people: it flags 14\% of
same-pipeline human frames, 13.6\% of yaw-matched ones and 17.5\% of FFHQ
images. Its large crops also cover the
robot's neck and cables (the human crops have the same geometry), so part of
the robot score may come from the robot's body. The two CelebA-Spoof
models are much less consistent: AENet flags 98.2\% of the panel frames but
none of the direct-capture or X2C frames, and MN3 flags only 18.3\% of the panel
frames, hardly more than the 13.9\% of yaw-matched humans it flags. Like the deepfake detectors, the PAD models do not separate the physical
robot from its synthesized video in a consistent way; AENet strongly reverses
the order (AUC 0.02). This is not a failure of PAD, which was never meant to
detect synthesis.

\subsection{Cost}
\label{sec:res-cost}

\begin{figure}[t]
\centering
\includegraphics[width=\columnwidth]{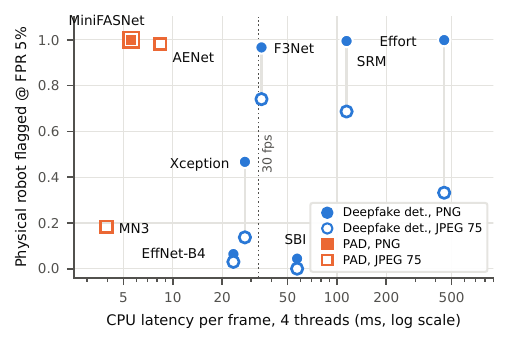}
\caption{CPU time per frame (4 threads) against how often the physical robot
is flagged at 5\% FPR. Circles: deepfake detectors; squares: PAD models.
Filled: PNG; open: JPEG 75. For AENet and MN3 only JPEG 75 is valid
(Sec.~\ref{sec:protocol}). Dotted line: 30\,fps.}
\label{fig:tradeoff}
\end{figure}

Figure~\ref{fig:tradeoff} puts cost next to the robot flag rate. On 4 CPU threads, only Xception (27.5\,ms) and
EfficientNet-B4 (23.4\,ms) of the deepfake detectors fit the 33\,ms budget of a
30\,fps stream, and neither flags more than half of the robot frames; F3Net is
just over (34.7\,ms), SBI needs 57.2\,ms, SRM 114.5\,ms and Effort, with 303M parameters, 450.6\,ms. The PAD models
need 3.9--8.4\,ms. MiniFASNet, with 0.87M parameters in total, is the only model
in this study that flags all physical robot frames, survives compression and runs far
within budget on a CPU.

\section{Discussion}
\label{sec:disc}

\paragraph{Today's robots are caught for the wrong reason.}
Several deepfake detectors flag the physical robot, but for the wrong reason.
With only the face visible, three of them score the physical robot higher than
its synthesized video, which a synthesis detector would not do. The robot
scores of most detectors sit close to the threshold, so compression pushes many of them below it.
And the detectors disagree widely on the same frames (4\% to almost 100\%);
a forgery trace they were trained to find would give much more agreement. This protection depends on how the
robot looks, and robot faces are moving toward human
appearance~\cite{sakamoto2007android,sato2022nikola,hu2024coexpression,hu2026lip}.

\textbf{H1.} As robot faces get more realistic in skin, eyes and fine motion,
their scores under current detectors will move toward the human distribution,
and the share flagged at a fixed low-FPR threshold will drop. H1 can be tested
by running our protocol on more realistic platforms, ideally with the same
identity filmed as a human, as a robot and as a deepfake.

\paragraph{Real-time interaction and physical presence.}
A RoboFake can hold a conversation like a live deepfake, but it renders
nothing. Challenge--response methods ask the caller to cover the face, turn or
press on the cheek~\cite{mittal2024gotcha}, and active probes shake the
camera~\cite{xie2024sfake}; both rely on the fake breaking when the scene
changes. A robot face does not break: occlusion, head turns and blur are
physically correct because they are physical.

\textbf{H2.} Challenge--response and active-probe defenses designed for live
deepfakes will pass a RoboFake that can perform the requested action.

Physical imitation of a living person is already studied in mobile liveness
detection, where motorized actuators mimic a user's involuntary hand
motion~\cite{gharib2026alive}. The same reasoning points to cues that should
work better against robots. A silicone face has no blood flow, so it should not
show the color changes that remote photoplethysmography picks up from human
skin~\cite{ciftci2020fakecatcher}. More generally, a model that asks whether a
face is alive, rather than whether an image was edited, targets what a RoboFake
actually lacks.

\textbf{H3.} Cues that require a living face will separate robots from humans
more reliably than forgery cues, and will be harder for an attacker to fake.

Our PAD results give first, mixed support for H3; testing H2 and H3 properly
needs video of a robot performing challenge actions, with temporal and
physiological cues.

\paragraph{Why PAD suits today's robots better.}
Two things in our data explain why the best PAD model detects the current robot
more reliably than every deepfake detector, despite its higher false-alarm rate
on processed human video. The first is what the models were trained to reject. Deepfake detectors learn traces of editing, which a robot
does not have, so their response to it is a side effect of the face looking
unusual. PAD models learn to reject non-living faces in front of the camera,
including masks and fake heads, and a robot face is much closer to these. The
second is the margin. MiniFASNet gives the robot a mean attack score of 0.97
against a threshold of 0.27, and AENet (JPEG 75) gives 0.78 against 0.15, so the
robot lies far from the decision boundary. For the deepfake detectors it lies
close to it: only F3Net (0.93 vs.\ 0.76) and Effort (0.83 vs.\ 0.70) are
clearly above; the other four are at most 0.03 above the threshold, or below. Under JPEG 75 the deepfake thresholds rise and Effort's mean robot
score (0.83) falls below its threshold (0.85), while MiniFASNet keeps a gap of
0.67, enough to survive compression. We think the PAD margin comes from cues that do not need fine pixel
detail, such as the color and shading of the skin material and, for MiniFASNet,
the robot's neck and cables in its wide crop, whereas deepfake detectors rely
more on fine detail that compression removes. These cues are also why we expect
the advantage to shrink: a robot with skin-like material and hidden mechanics
would remove them, and two of the three PAD models already miss the robot in
some captures.

\paragraph{Detection has to run at the edge.}
A RoboFake met live has to be caught on the device that holds the camera,
within about 33\,ms per frame. Speed is not the hard part (Sec.~\ref{sec:res-cost});
what is missing is a model this cheap that keeps false alarms on real people
low and does not depend on how the video was captured.

\section{Limitations and Ethics}
\label{sec:limits}

Our physical data come from one robot whose face is clearly not human, so H1
remains untested. The robot clips come from composite videos encoded several
times before we saw them; our same-pipeline control matches the face size and
the number of encoding rounds, but not the exact settings. The robot data
contain 13 composite videos of only three sentences and one direct-capture
clip; we resample whole composites, but with so few sentences our intervals may
still be too narrow. We tested only
frame-level models with public checkpoints, used a fixed face box for masking
and Grad-CAM, and checked the PAD models on only one PAD test set.

We did not build, modify or operate any robot, and we did not imitate any real
person. All data are public and used under their licenses. No identifiable human
face appears in the paper, and we describe how a RoboFake could be built only at
the level needed to motivate defenses.

\section{Conclusion}
\label{sec:conclusion}

We introduced RoboFake, a presentation attack by a physical humanoid face
robot. A small liveness model detects the physical robot reliably but has a high BPCER
on processed human video; other PAD models and the deepfake detectors give
inconsistent results that depend on the model, the capture and compression.
Today's models catch today's robots because the faces do not look human,
which will not last as robot faces become more realistic.

\bibliographystyle{IEEEtran}
\bibliography{refs}

@article{hu2026lip,
  author       = {Yuhang Hu and
                  Jiong Lin and
                  Judah A. Goldfeder and
                  Philippe Martin Wyder and
                  Yifeng Cao and
                  Steven Tian and
                  Yunzhe Wang and
                  Jingran Wang and
                  Mengmeng Wang and
                  Jie Zeng and
                  Cameron Mehlman and
                  Yingke Wang and
                  Delin Zeng and
                  Boyuan Chen and
                  Hod Lipson},
  title        = {Learning realistic lip motions for humanoid face robots},
  journal      = {Sci. Robotics},
  volume       = {11},
  number       = {110},
  year         = {2026}
}

@misc{hu2026lipdata,
  author       = {Hu, Yuhang and others},
  title        = {Data for: Learning realistic lip motions for humanoid face robots},
  howpublished = {Dryad, \url{https://datadryad.org/dataset/doi:10.5061/dryad.j6q573nrc}},
  year         = {2026}
}

@article{hu2024coexpression,
  author       = {Yuhang Hu and
                  Boyuan Chen and
                  Jiong Lin and
                  Yunzhe Wang and
                  Yingke Wang and
                  Cameron Mehlman and
                  Hod Lipson},
  title        = {Human-robot facial coexpression},
  journal      = {Sci. Robotics},
  volume       = {9},
  number       = {88},
  year         = {2024}
}

@article{li2025x2c,
  author  = {Li, Peizhen and Cao, Longbing and Wu, Xiao-Ming and Yang, Runze and Yu, Xiaohan},
  title   = {{X2C}: A Dataset Featuring Nuanced Facial Expressions for Realistic Humanoid Imitation},
  journal = {arXiv preprint arXiv:2505.11146},
  year    = {2025}
}

@inproceedings{zhu2025rena,
  author       = {Yongtong Zhu and
                  Lei Li and
                  Iggy Qian and
                  Wenbin Zhou and
                  Ye Yuan and
                  Qingdu Li and
                  Na Liu and
                  Jianwei Zhang},
  title        = {Awakening Facial Emotional Expressions in Human-Robot},
  booktitle    = {{IEEE/RSJ} International Conference on Intelligent Robots and Systems,
                  {IROS}},
  pages        = {21511--21518},
  publisher    = {{IEEE}},
  year         = {2025}
}

@article{sato2022nikola,
  author  = {Sato, Wataru  and Namba, Shushi  and Yang, Dongsheng  and Nishida, Shin’ya  and Ishi, Carlos  and Minato, Takashi},
  title   = {An Android for Emotional Interaction: Spatiotemporal Validation of Its Facial Expressions},
  journal = {Frontiers in Psychology},
  volume  = {12},
  pages   = {800657},
  year    = {2022}
}

@inproceedings{sakamoto2007android,
  author       = {Daisuke Sakamoto and
                  Takayuki Kanda and
                  Tetsuo Ono and
                  Hiroshi Ishiguro and
                  Norihiro Hagita},
  title        = {Android as a telecommunication medium with a human-like presence},
  booktitle    = {Proceedings of the Second {ACM} {SIGCHI/SIGART} Conference on Human-Robot
                  Interaction, {HRI}},
  pages        = {193--200},
  publisher    = {{ACM}},
  year         = {2007}
}

@article{mori2012uncanny,
  author       = {Masahiro Mori and
                  Karl F. MacDorman and
                  Norri Kageki},
  title        = {The Uncanny Valley [From the Field]},
  journal      = {IEEE Robotics \& Automation Magazine},
  volume       = {19},
  number       = {2},
  pages        = {98--100},
  year         = {2012}
}

@article{mathur2016navigating,
    title = {Navigating a social world with robot partners: A quantitative cartography of the Uncanny Valley},
    journal = {Cognition},
    volume = {146},
    pages = {22-32},
    year = {2016},
    author = {Maya B. Mathur and David B. Reichling}
}

@inproceedings{
    kim2026irobot,
    title={{I-Robot}: Identifying Robotic and Human Motion in Humanoids},
    author={Taehoon Kim and Jongwook Choi and Haeun Noh and Junyeup Hwang and Jongwon Choi},
    booktitle={Trustworthy AI for Good (AI4GOOD) Workshop @ ICML},
    year={2026}
}

@inproceedings{li2026motionturing,
  author       = {Mingzhe Li and
                  Mengyin Liu and
                  Zekai Wu and
                  Xincheng Lin and
                  Junsheng Zhang and
                  Ming Yan and
                  Zengye Xie and
                  Changwang Zhang and
                  Chenglu Wen and
                  Lan Xu and
                  Siqi Shen and
                  Cheng Wang},
  title        = {Towards Motion {Turing} Test: Evaluating Human-Likeness in Humanoid
                  Robots},
  booktitle    = {{IEEE/CVF} Conference on Computer Vision and Pattern Recognition,
                  {CVPR}},
  pages        = {16486--16498},
  year         = {2026}
}

@article{george2019wmca,
  author       = {Anjith George and
                  Zohreh Mostaani and
                  David Geissenbuhler and
                  Olegs Nikisins and
                  Andr{\'{e}} Anjos and
                  S{\'{e}}bastien Marcel},
  title        = {Biometric Face Presentation Attack Detection With Multi-Channel Convolutional
                  Neural Network},
  journal      = {{IEEE} Transactions on Information Forensics and Security},
  volume       = {15},
  pages        = {42--55},
  year         = {2020}
}

@inproceedings{liu2021hifimask,
  author       = {Ajian Liu and
                  Chenxu Zhao and
                  Zitong Yu and
                  Anyang Su and
                  Xing Liu and
                  Zijian Kong and
                  Jun Wan and
                  Sergio Escalera and
                  Hugo Jair Escalante and
                  Zhen Lei and
                  Guodong Guo},
  title        = {{3D} High-Fidelity Mask Face Presentation Attack Detection Challenge},
  booktitle    = {{IEEE/CVF} International Conference on Computer Vision Workshops,
                  {ICCVW}},
  pages        = {814--823},
  year         = {2021}
}

@article{yu2023fassurvey,
  author       = {Zitong Yu and
                  Yunxiao Qin and
                  Xiaobai Li and
                  Chenxu Zhao and
                  Zhen Lei and
                  Guoying Zhao},
  title        = {Deep Learning for Face Anti-Spoofing: {A} Survey},
  journal      = {{IEEE} Transactions on Pattern Analysis \& Machine Intelligence},
  volume       = {45},
  pages        = {5609--5631},
  year         = {2023}
}

@inproceedings{rossler2019ffpp,
  author       = {Andreas R{\"{o}}ssler and
                  Davide Cozzolino and
                  Luisa Verdoliva and
                  Christian Riess and
                  Justus Thies and
                  Matthias Nie{\ss}ner},
  title        = {{FaceForensics++}: Learning to Detect Manipulated Facial Images},
  booktitle    = {2019 {IEEE/CVF} International Conference on Computer Vision, {ICCV}},
  pages        = {1--11},
  publisher    = {{IEEE}},
  year         = {2019}
}

@inproceedings{yan2023deepfakebench,
  author       = {Zhiyuan Yan and
                  Yong Zhang and
                  Xinhang Yuan and
                  Siwei Lyu and
                  Baoyuan Wu},
  title        = {{DeepfakeBench}: {A} Comprehensive Benchmark of Deepfake Detection},
  booktitle    = {Advances in Neural Information Processing Systems 36: Annual Conference
                  on Neural Information Processing Systems 2023, NeurIPS},
  year         = {2023}
}

@inproceedings{chollet2017xception,
  author       = {Fran{\c{c}}ois Chollet},
  title        = {Xception: Deep Learning with Depthwise Separable Convolutions},
  booktitle    = {2017 {IEEE} Conference on Computer Vision and Pattern Recognition,
                  {CVPR}},
  pages        = {1800--1807},
  publisher    = {{IEEE} Computer Society},
  year         = {2017}
}

@inproceedings{tan2019efficientnet,
  author       = {Mingxing Tan and
                  Quoc V. Le},
  title        = {{EfficientNet}: Rethinking Model Scaling for Convolutional Neural Networks},
  booktitle    = {Proceedings of the 36th International Conference on Machine Learning,
                  {ICML}},
  series       = {Proceedings of Machine Learning Research},
  pages        = {6105--6114},
  year         = {2019}
}

@inproceedings{qian2020f3net,
  author       = {Yuyang Qian and
                  Guojun Yin and
                  Lu Sheng and
                  Zixuan Chen and
                  Jing Shao},
  title        = {Thinking in Frequency: Face Forgery Detection by Mining Frequency-Aware
                  Clues},
  booktitle    = {Computer Vision - {ECCV} 2020 - 16th European Conference, Glasgow,
                  UK, August 23-28, 2020, Proceedings, Part {XII}},
  series       = {Lecture Notes in Computer Science},
  pages        = {86--103},
  year         = {2020}
}

@inproceedings{luo2021srm,
  author       = {Yuchen Luo and
                  Yong Zhang and
                  Junchi Yan and
                  Wei Liu},
  title        = {Generalizing Face Forgery Detection With High-Frequency Features},
  booktitle    = {{IEEE} Conference on Computer Vision and Pattern Recognition, {CVPR}},
  pages        = {16317--16326},
  publisher    = {Computer Vision Foundation / {IEEE}},
  year         = {2021}
}

@inproceedings{shiohara2022sbi,
  author       = {Kaede Shiohara and
                  Toshihiko Yamasaki},
  title        = {Detecting Deepfakes with Self-Blended Images},
  booktitle    = {{IEEE/CVF} Conference on Computer Vision and Pattern Recognition,
                  {CVPR}},
  pages        = {18699--18708},
  publisher    = {{IEEE}},
  year         = {2022}
}

@inproceedings{yan2025effort,
  author       = {Zhiyuan Yan and
                  Jiangming Wang and
                  Peng Jin and
                  Ke{-}Yue Zhang and
                  Chengchun Liu and
                  Shen Chen and
                  Taiping Yao and
                  Shouhong Ding and
                  Baoyuan Wu and
                  Li Yuan},
  title        = {Orthogonal Subspace Decomposition for Generalizable {AI}-Generated Image
                  Detection},
  booktitle    = {Forty-second International Conference on Machine Learning, {ICML}},
  series       = {Proceedings of Machine Learning Research},
  year         = {2025}
}

@inproceedings{radford2021clip,
  author       = {Alec Radford and
                  Jong Wook Kim and
                  Chris Hallacy and
                  Aditya Ramesh and
                  Gabriel Goh and
                  Sandhini Agarwal and
                  Girish Sastry and
                  Amanda Askell and
                  Pamela Mishkin and
                  Jack Clark and
                  Gretchen Krueger and
                  Ilya Sutskever},
  title        = {Learning Transferable Visual Models From Natural Language Supervision},
  booktitle    = {Proceedings of the 38th International Conference on Machine Learning,
                  {ICML}},
  series       = {Proceedings of Machine Learning Research},
  pages        = {8748--8763},
  year         = {2021}
}

@inproceedings{haliassos2021lips,
  author       = {Alexandros Haliassos and
                  Konstantinos Vougioukas and
                  Stavros Petridis and
                  Maja Pantic},
  title        = {Lips Don't Lie: {A} Generalisable and Robust Approach To Face Forgery
                  Detection},
  booktitle    = {{IEEE} Conference on Computer Vision and Pattern Recognition, {CVPR}},
  pages        = {5039--5049},
  publisher    = {Computer Vision Foundation / {IEEE}},
  year         = {2021}
}

@article{ciftci2020fakecatcher,
  title={{FakeCatcher}: Detection of synthetic portrait videos using biological signals},
  author={Ciftci, Umur Aybars and Demir, Ilke and Yin, Lijun},
  journal={{IEEE} Transactions on Pattern Analysis and Machine Intelligence},
  year={2020},
  publisher={IEEE}
}

@inproceedings{mittal2024gotcha,
  author       = {Govind Mittal and
                  Chinmay Hegde and
                  Nasir D. Memon},
  title        = {{GOTCHA}: Real-Time Video Deepfake Detection via Challenge-Response},
  booktitle    = {9th {IEEE} European Symposium on Security and Privacy, EuroS{\&}P},
  pages        = {1--20},
  publisher    = {{IEEE}},
  year         = {2024}
}

@misc{heygem,
  author       = {{Duix}},
  title        = {{HeyGem} (now {Duix-Avatar}): Open-source {AI} avatar toolkit},
  howpublished = {\url{https://github.com/duixcom/Duix-Avatar}},
  note         = {Accessed Oct. 2026},
  year         = {2025}
}

@article{wu2023yunet,
  author       = {Wei Wu and
                  Hanyang Peng and
                  Shiqi Yu},
  title        = {{YuNet}: {A} Tiny Millisecond-level Face Detector},
  journal      = {Machine Intelligence Research},
  volume       = {20},
  pages        = {656--665},
  year         = {2023}
}

@inproceedings{karras2019stylegan,
  author       = {Tero Karras and
                  Samuli Laine and
                  Timo Aila},
  title        = {A Style-Based Generator Architecture for Generative Adversarial Networks},
  booktitle    = {{IEEE} Conference on Computer Vision and Pattern Recognition, {CVPR}},
  pages        = {4401--4410},
  publisher    = {Computer Vision Foundation / {IEEE}},
  year         = {2019}
}

@inproceedings{selvaraju2017gradcam,
  author       = {Ramprasaath R. Selvaraju and
                  Michael Cogswell and
                  Abhishek Das and
                  Ramakrishna Vedantam and
                  Devi Parikh and
                  Dhruv Batra},
  title        = {{Grad-CAM}: Visual Explanations from Deep Networks via Gradient-Based
                  Localization},
  booktitle    = {{IEEE} International Conference on Computer Vision, {ICCV}},
  pages        = {618--626},
  publisher    = {{IEEE} Computer Society},
  year         = {2017}
}

@article{xie2024sfake,
  title={Shaking the fake: Detecting deepfake videos in real time via active probes},
  author={Xie, Zhixin and Luo, Jun},
  journal={arXiv preprint arXiv:2409.10889},
  year={2024}
}

@article{laionmobile2026,
  title={{LAION-Mobile}: Evaluating Deepfake Detectors On One Million Smartphone Photos},
  author={von Stryk, Achim and Keuper, Janis},
  journal={arXiv preprint arXiv:2609.11134},
  year={2026}
}

@article{gharib2026alive,
  title={{A-Live}: Passive Liveness Detection via Neuromuscular Micro-Motion Signatures on Commodity Sensors},
  author={Gharib, Mohammed and Burns, Sam and Zizi, Martin},
  journal={arXiv preprint arXiv:2606.05126},
  year={2026}
}

@inproceedings{liu2019siwm,
  author       = {Yaojie Liu and
                  Joel Stehouwer and
                  Amin Jourabloo and
                  Xiaoming Liu},
  title        = {Deep Tree Learning for Zero-Shot Face Anti-Spoofing},
  booktitle    = {{IEEE} Conference on Computer Vision and Pattern Recognition, {CVPR}},
  pages        = {4680--4689},
  publisher    = {Computer Vision Foundation / {IEEE}},
  year         = {2019}
}

@inproceedings{zhang2020celebaspoof,
  author       = {Yuanhan Zhang and
                  Zhenfei Yin and
                  Yidong Li and
                  Guojun Yin and
                  Junjie Yan and
                  Jing Shao and
                  Ziwei Liu},
  title        = {{CelebA-Spoof}: Large-Scale Face Anti-spoofing Dataset with Rich Annotations},
  booktitle    = {European conference on computer vision},
  series       = {Lecture Notes in Computer Science},
  volume       = {12357},
  pages        = {70--85},
  publisher    = {Springer},
  year         = {2020}
}

@misc{silentface,
  author       = {{Minivision AI}},
  title        = {Silent-Face-Anti-Spoofing},
  howpublished = {\url{https://github.com/minivision-ai/Silent-Face-Anti-Spoofing}},
  year         = {2020}
}

@misc{prokofiev2020mn3,
  author       = {Prokofiev, Kirill},
  title        = {Light-weight face anti-spoofing},
  howpublished = {\url{https://github.com/kprokofi/light-weight-face-anti-spoofing}},
  year         = {2020}
}

@misc{iso30107,
  author       = {{ISO/IEC JTC1 SC37}},
  title        = {Information Technology -- Biometric Presentation Attack Detection -- Part 3: Testing and Reporting},
  howpublished = {ISO/IEC 30107-3},
  year         = {2023}
}

\end{document}